# DATA SPECTROSCOPY: EIGENSPACES OF CONVOLUTION OPERATORS AND CLUSTERING

### By Tao Shi[1], Mikhail Belkin[2] and Bin Yu[3]

*Ohio State University, Ohio State University and
University of California, Berkeley*

This paper focuses on obtaining clustering information about a distribution from its i.i.d. samples. We develop theoretical results to understand and use clustering information contained in the eigenvectors of data adjacency matrices based on a radial kernel function with a sufficiently fast tail decay. In particular, we provide population analyses to gain insights into which eigenvectors should be used and when the clustering information for the distribution can be recovered from the sample. We learn that a fixed number of top eigenvectors might at the same time contain redundant clustering information and miss relevant clustering information. We use this insight to design the *data spectroscopic* clustering (DaSpec) algorithm that utilizes properly selected eigenvectors to determine the number of clusters automatically and to group the data accordingly. Our findings extend the intuitions underlying existing spectral techniques such as spectral clustering and Kernel Principal Components Analysis, and provide new understanding into their usability and modes of failure. Simulation studies and experiments on real-world data are conducted to show the potential of our algorithm. In particular, DaSpec is found to handle unbalanced groups and recover clusters of different shapes better than the competing methods.

**1. Introduction.** Data clustering based on eigenvectors of a proximity or affinity matrix (or its normalized versions) has become popular in machine learning, computer vision and many other areas. Given data $x_1, \ldots, x_n \in \mathbb{R}^d$,

---

Received July 2008; revised March 2009.

[1]Supported in part by NASA Grant NNG06GD31G.

[2]Supported in part by NSF Early Career Award 0643916.

[3]Supported in part by NSF Grant DMS-06-05165, ARO Grant W911NF-05-1-0104, NSFC Grant 60628102, a grant from MSRA and a Guggenheim Fellowship in 2006.

this family of algorithms constructs an affinity matrix $(K_n)_{ij} = K(x_i, x_j)/n$ based on a kernel function, such as a Gaussian kernel $K(x, y) = e^{-\|x-y\|^2/(2\omega^2)}$. Clustering information is obtained by taking eigenvectors and eigenvalues of the matrix $K_n$ or the closely related graph Laplacian matrix $L_n = D_n - K_n$, where $D_n$ is a diagonal matrix with $(D_n)_{ii} = \sum_j (K_n)_{ij}$. The basic intuition is that when the data come from several clusters, distances between clusters are typically far larger than the distances within the same cluster, and thus $K_n$ and $L_n$ are (close to) block-diagonal matrices up to a permutation of the points. Eigenvectors of such block-diagonal matrices keep the same structure. For example, the few top eigenvectors of $L_n$ can be shown to be constant on each cluster, assuming infinite separation between clusters, allowing one to distinguish the clusters by looking for data points corresponding to the same or similar values of the eigenvectors.

In particular, we note the algorithm of Scott and Longuet-Higgins [13] who proposed to embed data into the space spanned by the top eigenvectors of $K_n$, normalize the data in that space and group data by investigating the block structure of inner product matrix of normalized data. Perona and Freeman [10] suggested to cluster the data into two groups by directly thresholding the top eigenvector of $K_n$.

Another important algorithm, the *normalized cut*, was proposed by Shi and Malik [14] in the context of image segmentation. It separates data into two groups by thresholding the second smallest generalized eigenvector of $L_n$. Assuming $k$ groups, Malik et al. [6] and Ng, Jordan and Weiss [8] suggested embedding the data into the span of the bottom $k$ eigenvectors of the normalized graph Laplacian[1] $I_n - D_n^{-1/2} K_n D_n^{-1/2}$ and applying the $k$-means algorithm to group the data in the embedding space. For further discussions on spectral clustering, we refer the reader to Weiss [20], Dhillon, Guan and Kulis [2] and von Luxburg [18]. An empirical comparison of various methods is provided in Verma and Meila [17]. A discussion of some limitations of spectral clustering can be found in Nadler and Galun [7]. A theoretical analysis of statistical consistency of different types of spectral clustering is provided in von Luxburg, Belkin and Bousquet [19].

Similarly to spectral clustering methods, Kernel Principal Component Analysis (Schölkopf, Smola and Müller [12]) and spectral dimensionality reduction (e.g., Belkin and Niyogi [1]) seek lower dimensional representations of the data by embedding them into the space spanned by the top eigenvectors of $K_n$ or the bottom eigenvectors of the normalized graph Laplacian with the expectation that this embedding keeps nonlinear structure of the data. Empirical observations have also been made that KPCA can sometimes capture clusters in the data. The concept of using eigenvectors of the kernel matrix is also closely connected to other kernel methods in the machine learning literature, notably Support Vector Machines (cf. Vapnik [16] and Schölkopf and Smola [11]), which can be viewed as fitting a linear classifier in the eigenspace of $K_n$.

---

[1] We assume here that the diagonal terms of $K_n$ are replaced by zeros.

Although empirical results and theoretical studies both suggest that the top eigenvectors contain clustering information, the effectiveness of these algorithms hinges heavily on the choice of the kernel and its parameters, the number of the top eigenvectors used, and the number of groups employed. As far as we know, there are no explicit theoretical results or practical guidelines on how to make these choices. Instead of tackling these questions regarding to particular data sets, it may be more fruitful to investigate them from a population point of view. Williams and Seeger [21] investigated the dependence of the spectrum of $K_n$ on the data density function and analyzed this dependence in the context of lower rank matrix approximations to the kernel matrix. To the best of our knowledge, this work was the first theoretical study of this dependence.

In this paper we aim to understand spectral clustering methods based on a population analysis. We concentrate on exploring the connections between the distribution $P$ and the eigenvalues and eigenfunctions of the distribution-dependent convolution operator,

$$(1.1) \qquad \mathcal{K}_P f(x) = \int K(x,y) f(y) \, dP(y).$$

The kernels we consider will be positive (semi-)definite radial kernels. Such kernels can be written as $K(x,y) = k(\|x - y\|)$, where $k : [0, \infty) \to [0, \infty)$ is a decreasing function. We will use kernels with sufficiently fast tail decay, such as the Gaussian kernel or the exponential kernel $K(x,y) = e^{-\|x-y\|/\omega}$. The connections found allow us to gain some insights into when and why these algorithms are expected to work well. In particular, we learn that a fixed number of top eigenvectors of the kernel matrix do not always contain all of the clustering information. In fact, when the clusters are not balanced and/or have different shapes, the top eigenvectors may be inadequate and redundant at the same time. That is, some of the top eigenvectors may correspond to the same cluster while missing other significant clusters. Consequently, we devise a clustering algorithm that selects only those eigenvectors which have clustering information not represented by the other eigenvectors already selected.

The rest of the paper is organized as follows. In Section 2, we cover the basic definitions, notation and mathematical facts about the distribution-dependent convolution operator and its spectrum. We point out the strong connection between $\mathcal{K}_P$ and its empirical version, the kernel matrix $K_n$, which allows us to approximate the spectrum of $\mathcal{K}_P$ given data.

In Section 3, we characterize the dependence of eigenfunctions of $\mathcal{K}_P$ on both the distribution $P$ and the kernel function $K(\cdot, \cdot)$. We show that the eigenfunctions of $\mathcal{K}_P$ decay to zero at the tails of the distribution $P$ and that their decay rates depends on both the tail decay rate of $P$ and that of the kernel $K(\cdot, \cdot)$. For distributions with only one high density component, we provide theoretical analysis. A discussion of three special cases can be found in the Appendix A. In the first two examples, the exact form of the eigenfunctions of $\mathcal{K}_P$ can be found; in the third, the distribution is concentrated on or around a curve in $\mathbb{R}^d$.

Further, we consider the case when the distribution $P$ contains several separate high-density components. Through classical results of the perturbation theory, we show that the top eigenfunctions of $\mathcal{K}_P$ are approximated by the top eigenfunctions of the corresponding operators defined on some of those components. However, not every component will contribute to the top few eigenfunctions of $\mathcal{K}_P$ as the eigenvalues are determined by the size and configuration of the corresponding component. Based on this key property, we show that the top eigenvectors of the kernel matrix may or may not preserve all clustering information, which explains some empirical observations of certain spectral clustering methods. A real-world high-dimensional dataset, the USPS postal code digit data, is also analyzed to illustrate this property.

In Section 4, we utilize our theoretical results to construct the *data spectroscopic* clustering (DaSpec) algorithm that estimates the number of groups data-dependently, assigns labels to each observation, and provides a classification rule for unobserved data, all based on the same eigen decomposition. Data-dependent choices of algorithm parameters are also discussed. In Section 5, the proposed DaSpec algorithm is tested on two simulations against commonly used $k$-means and spectral clustering algorithms. In both situations, the DaSpec algorithm provides favorable results even when the other two algorithms are provided with the number of groups in advance. Section 6 contains conclusions and discussion.

## 2. Notation and mathematical preliminaries.

2.1. *Distribution-dependent convolution operator.* Given a probability distribution $P$ on $\mathbb{R}^d$, we define $L^2_P(\mathbb{R}^d)$ to be the space of square integrable functions, $f \in L^2_P(\mathbb{R}^d)$ if $\int f^2 \, dP < \infty$, and the space is equipped with an inner product $\langle f, g \rangle = \int fg \, dP$. Given a kernel (symmetric function of two variables) $K(x, y) : \mathbb{R}^d \times \mathbb{R}^d \to \mathbb{R}$, (1.1) defines the corresponding integral operator $\mathcal{K}_P$. Recall that an eigenfunction $\phi : \mathbb{R}^d \mapsto \mathbb{R}$ and the corresponding eigenvalue $\lambda$ of $\mathcal{K}_P$ are defined by the following equations:

$$(2.1) \qquad\qquad \mathcal{K}_P \phi = \lambda \phi,$$

and the constraint $\int \phi^2 \, dP = 1$. If the kernel satisfies the condition

$$(2.2) \qquad \iint K^2(x,y) \, dP(x) \, dP(y) < \infty,$$

the corresponding operator $\mathcal{K}_P$ is a trace class operator, which, in turn, implies that it is compact and has a discrete spectrum.

In this paper, we will only consider the case when a positive semi-definite kernel $K(x,y)$ and a distribution $P$ generate a trace class operator $\mathcal{K}_P$, so that it has only countable nonnegative eigenvalues $\lambda_0 \geq \lambda_1 \geq \lambda_2 \geq \cdots \geq 0$. Moreover, there is a corresponding orthonormal basis in $L_p^2$ of eigenfunctions $\phi_i$ satisfying (2.1). The dependence of the eigenvalues and eigenfunctions of $\mathcal{K}_P$ on $P$ will be one of the main foci of our paper. We note that an eigenfunction $\phi$ is uniquely defined not only on the support of $P$, but on every point $x \in \mathbb{R}^d$ through $\phi(x) = \frac{1}{\lambda} \int K(x,y)\phi(y) \, dP(y)$, assuming that the kernel function $K$ is defined everywhere on $\mathbb{R}^d \times \mathbb{R}^d$.

2.2. *Kernel matrix.* Let $x_1, \ldots, x_n$ be an i.i.d. sample drawn from distribution $P$. The corresponding empirical operator $\mathcal{K}_{P_n}$ is defined as

$$\mathcal{K}_{P_n} f(x) = \int K(x,y) f(y) \, dP_n(y) = \frac{1}{n} \sum_{i=1}^{n} K(x,x_i) f(x_i).$$

This operator is closely related to the $n \times n$ kernel matrix $K_n$, where

$$(K_n)_{ij} = K(x_i, x_j)/n.$$

Specifically, the eigenvalues of $\mathcal{K}_{P_n}$ are the same as those of $K_n$ and an eigenfunction $\phi$, with an eigenvalue $\lambda \neq 0$ of $\mathcal{K}_{P_n}$, is connected with the corresponding eigenvector $\mathbf{v} = [v_1, v_2, \ldots, v_n]'$ of $K_n$ by

$$\phi(x) = \frac{1}{n\lambda} \sum_{i=1}^{n} K(x,x_i) v_i \qquad \forall x \in \mathbb{R}^d.$$

It is easy to verify that $\mathcal{K}_{P_n} \phi = \lambda \phi$. Thus values of $\phi$ at locations $x_1, \ldots, x_n$ coincide with the corresponding entries of the eigenvector $\mathbf{v}$. However, unlike $\mathbf{v}$, $\phi$ is defined everywhere in $\mathbb{R}^d$. For the spectrum of $\mathcal{K}_{P_n}$ and $K_n$, the only difference is that the spectrum of $\mathcal{K}_{P_n}$ contains 0 with infinite multiplicity. The corresponding eigenspace includes all functions vanishing on the sample points.

It is well known that, under mild conditions and when $d$ is fixed, the eigenvectors and eigenvalues of $K_n$ converge to eigenfunctions and eigenvalues of $\mathcal{K}_P$ as $n \to \infty$ (e.g., Koltchinskii and Giné [4]). Therefore, we expect the properties of the top eigenfunctions and eigenvalues of $\mathcal{K}_P$ also hold for $K_n$, assuming that $n$ is reasonably large.

**3. Spectral properties of $\mathcal{K}_P$.** In this section, we study the spectral properties of $\mathcal{K}_P$ and their connection to the data generating distribution $P$. We start with several basic properties of the top spectrum of $\mathcal{K}_P$ and then investigate the case when the distribution $P$ is a mixture of several high-density components.

3.1. *Basic spectral properties of $\mathcal{K}_P$.* Through Theorem 1 and its corollary, we obtain an important property of the eigenfunctions of $\mathcal{K}_P$, that is, these eigenfunctions decay fast when away from the majority of masses of the distribution if the tails of $K$ and $P$ have a fast decay. A second theorem offers the important property that the top eigenfunction has no sign change and multiplicity one. (Three detailed examples are provided in Appendix A to illustrate these two important properties.)

THEOREM 1 (Tail decay property of eigenfunctions). *An eigenfunction $\phi$ with the corresponding eigenvalue $\lambda > 0$ of $\mathcal{K}_P$ satisfies*

$$|\phi(x)| \leq \frac{1}{\lambda}\sqrt{\int [K(x,y)]^2 \, dP(y)}.$$

PROOF. By the Cauchy–Schwarz inequality and the definition of eigenfunction (2.1), we see that

$$\lambda|\phi(x)| = \left|\int K(x,y)\phi(y) \, dP(y)\right| \leq \int K(x,y)|\phi(y)| \, dP(y)$$

$$\leq \sqrt{\int [K(x,y)]^2 \, dP(y)}\sqrt{\int [\phi(y)]^2 \, dP(y)} = \sqrt{\int [K(x,y)]^2 \, dP(y)}.$$

The conclusion follows. □

We see that the "tails" of eigenfunctions of $\mathcal{K}_P$ decay to zero and that the decay rate depends on the tail behaviors of both the kernel $K$ and the distribution $P$. This observation will be useful to separate high-density areas in the case of $P$ having several components. Actually, we have the following corollary immediately:

COROLLARY 1. *Let $K(x,y) = k(\|x-y\|)$ and $k(\cdot)$ being nonincreasing. Assume that $P$ is supported on a compact set $D \subset \mathbb{R}^d$. Then*

$$|\phi(x)| \leq \frac{k(\mathrm{dist}(x,D))}{\lambda},$$

*where $\mathrm{dist}(x,D) = \inf_{y \in D}\|x-y\|$.*

The proof follows from Theorem 1 and the fact that $k(\cdot)$ is a nonincreasing function. And now we give an important property of the top (corresponding to the largest eigenvalue) eigenfunction.

THEOREM 2 (Top eigenfunction). *Let $K(x,y)$ be a positive semi-definite kernel with full support on $\mathbb{R}^d$. The top eigenfunction $\phi_0(x)$ of the convolution operator $\mathcal{K}_P$:*

1. *is the only eigenfunction with no sign change on $\mathbb{R}^d$;*
2. *has multiplicity one;*
3. *is nonzero on the support of $P$.*

The proof is given in Appendix B and these properties will be used later when we propose our clustering algorithm in Section 4.

3.2. *An example: top eigenfunctions of $\mathcal{K}_P$ for mixture distributions.* We now study the spectrum of $\mathcal{K}_P$ defined on a mixture distribution

$$(3.1) \qquad P = \sum_{g=1}^{G} \pi^g P^g,$$

which is a commonly used model in clustering and classification. To reduce notation confusion, we use *italicized superscript 1, 2, . . . , g, . . . , G* as the index of the mixing component and ordinary superscript for the power of a number. For each mixing component $P^g$, we define the corresponding operator $\mathcal{K}_{P^g}$ as

$$\mathcal{K}_{P^g} f(x) = \int K(x,y) f(y) \, dP^g(y).$$

We start by a mixture Gaussian example given in Figure 1. Gaussian kernel matrices $K_n$, $K_n^1$ and $K_n^2$ ($\omega = 0.3$) are constructed on three batches of 1000 i.i.d. samples from each of the three distributions: $0.5N(2, 1^2) + 0.5N(-2, 1^2)$, $N(2, 1^2)$ and $N(-2, 1^2)$. We observe that the top eigenvectors of $K_n$ are nearly identical to the top eigenvectors of $K_n^1$ or $K_n^2$.

From the point of view of the operator theory, it is easy to understand this phenomenon: *with a properly chosen kernel, the top eigenfunctions of an operator defined on each mixing component are approximate eigenfunctions of the operator defined on the mixture distribution.* To be explicit, let us consider the Gaussian convolution operator $\mathcal{K}_P$ defined by $P = \pi^1 P^1 + \pi^2 P^2$, with Gaussian components $P^1 = N(\mu^1, [\sigma^1]^2)$ and $P^2 = N(\mu^2, [\sigma^2]^2)$ and the Gaussian kernel $K(x,y)$ with bandwidth $\omega$. Due to the linearity of convolution operators, $\mathcal{K}_P = \pi^1 \mathcal{K}_{P^1} + \pi^2 \mathcal{K}_{P^2}$.

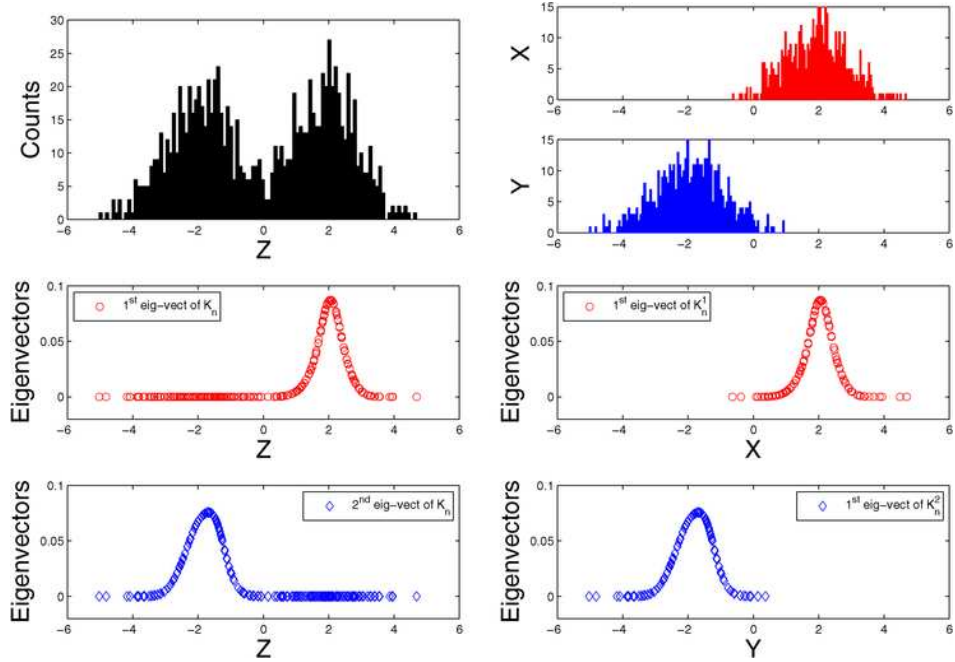

FIG. 1. *Eigenvectors of a Gaussian kernel matrix ($\omega = 0.3$) of 1000 data sampled from a mixture Gaussian distribution $0.5N(2, 1^2) + 0.5N(-2, 1^2)$. Left panels: histogram of the data (top), first eigenvector of $K_n$ (middle) and second eigenvector of $K_n$ (bottom). Right panels: histograms of data from each component (top), first eigenvector of $K_n^1$ (middle) and first eigenvector of $K_n^2$ (bottom).*

Consider an eigenfunction $\phi^1(x)$ of $\mathcal{K}_{P^1}$ with the corresponding eigenvalue $\lambda^1$, $\mathcal{K}_{P^1}\phi^1(x) = \lambda^1\phi^1(x)$. We have

$$\mathcal{K}_P\phi^1(x) = \pi^1\lambda^1\phi^1(x) + \pi^2 \int K(x, y)\phi^1(y)\, dP^2(y).$$

As shown in Proposition 1 in Appendix A, in the Gaussian case, $\phi^1(x)$ is centered at $\mu^1$ and its tail decays exponentially. Therefore, assuming enough separation between $\mu^1$ and $\mu^2$, $\pi^2 \int K(x, y)\phi^1(y)\, dP^2(y)$ is close to 0 everywhere, and hence $\phi^1(x)$ is an approximate eigenfunction of $\mathcal{K}_P$. In the next section, we will show that a similar approximation holds for general mixture distributions whose components may not be Gaussian distributions.

3.3. *Perturbation analysis.* For $\mathcal{K}_P$ defined by a mixture distribution (3.1) and a positive semi-definite kernel $K(\cdot, \cdot)$, we now study the connection between its top eigenvalues and eigenfunctions and those of each $\mathcal{K}_{P^g}$. Without loss of generality, let us consider a mixture of two components. We state the following theorem regarding the top eigenvalue $\lambda_0$ of $\mathcal{K}_P$.

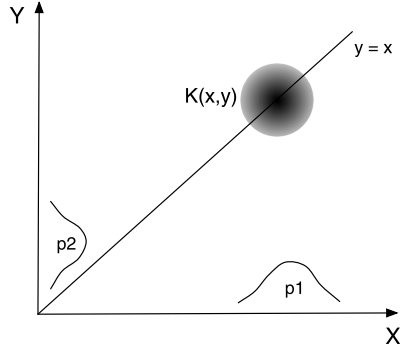

Fig. 2. *Illustration of separation condition (3.2) in Theorem 3.*

THEOREM 3 (Top eigenvalue of mixture distribution). *Let $P = \pi^1 P^1 + \pi^2 P^2$ be a mixture distribution on $\mathbb{R}^d$ with $\pi^1 + \pi^2 = 1$. Given a positive semi-definite kernel $K$, denote the top eigenvalue of $\mathcal{K}_P$, $\mathcal{K}_{P^1}$ and $\mathcal{K}_{P^2}$ as $\lambda_0$, $\lambda_0^1$ and $\lambda_0^2$, respectively. Then $\lambda_0$ satisfies*

$$\max(\pi^1 \lambda_0^1, \pi^2 \lambda_0^2) \leq \lambda_0 \leq \max(\pi^1 \lambda_0^1, \pi^2 \lambda_0^2) + r,$$

*where*

$$(3.2) \qquad r = \left( \pi^1 \pi^2 \iint [K(x,y)]^2 \, dP^1(x) \, dP^2(y) \right)^{1/2}.$$

The proof is given in Appendix B. As illustrated in Figure 2, the value of $r$ in (3.2) is small when $P^1$ and $P^2$ do not overlap much. Meanwhile, the size of $r$ is also affected by how fast $K(x,y)$ approaches zero as $\|x-y\|$ increases. When $r$ is small, the top eigenvalue of $\mathcal{K}_P$ is close to the larger one of $\pi^1 \lambda_0^1$ and $\pi^2 \lambda_0^2$. Without loss of generality, we assume $\pi^1 \lambda_0^1 > \pi^2 \lambda_0^2$ in the rest of this section.

The next lemma is a general perturbation result for the eigenfunctions of $\mathcal{K}_P$. The empirical (matrix) version of this lemma appeared in Diaconis, Goel and Holmes [3] and more general results can be traced back to Parlett [9].

LEMMA 1. *Consider an operator $\mathcal{K}_P$ with the discrete spectrum $\lambda_0 \geq \lambda_1 \geq \cdots$. If*

$$\|\mathcal{K}_P f - \lambda f\|_{L_P^2} \leq \varepsilon$$

*for some $\lambda$, $\varepsilon > 0$, and $f \in L_P^2$, then $\mathcal{K}_P$ has an eigenvalue $\lambda_k$ such that $|\lambda_k - \lambda| \leq \varepsilon$. If we further assume that $s = \min_{i : \lambda_i \neq \lambda_k} |\lambda_i - \lambda_k| > \varepsilon$, then $\mathcal{K}_P$ has an eigenfunction $f_k$ corresponding to $\lambda_k$ such that $\|f - f_k\|_{L_P^2} \leq \frac{\varepsilon}{s - \varepsilon}$.*

The lemma shows that a constant $\lambda$ must be "close" to an eigenvalue of $\mathcal{K}_P$ if the operator "almost" projects a function $f$ to $\lambda f$. Moreover, the function $f$ must be "close" to an eigenfunction of $\mathcal{K}_P$ if the distance between $\mathcal{K}_P f$ and $\lambda f$ is smaller than the eigen-gaps between $\lambda_k$ and other eigenvalues of $\mathcal{K}_P$. We are now in a position to state the perturbation result for the top eigenfunction of $\mathcal{K}_P$. Given the facts that $|\lambda_0 - \pi^1 \lambda_0^1| \leq r$ and

$$\mathcal{K}_P \phi_0^1 = \pi^1 \mathcal{K}_{P^1} \phi_0^1 + \pi^2 \mathcal{K}_{P^2} \phi_0^1 = (\pi^1 \lambda_0^1) \phi_0^1 + \pi^2 \mathcal{K}_{P^2} \phi_0^1,$$

Lemma 1 indicates that $\phi_0^1$ is close to $\phi_0$ if $\|\pi^2 \mathcal{K}_{P^2} \phi_0^1\|_{L_P^2}$ is small enough. To be explicit, we formulate the following corollary.

COROLLARY 2 (Top eigenfunction of mixture distribution). *Let* $P = \pi^1 P^1 + \pi^2 P^2$ *be a mixture distribution on* $\mathbb{R}^d$ *with* $\pi^1 + \pi^2 = 1$. *Given a semi-positive definite kernel* $K(\cdot, \cdot)$, *we denote the top eigenvalues of* $\mathcal{K}_{P^1}$ *and* $\mathcal{K}_{P^2}$ *as* $\lambda_0^1$ *and* $\lambda_0^2$, *respectively (assuming* $\pi^1 \lambda_0^1 > \pi^2 \lambda_0^2$) *and define* $t = \lambda_0 - \lambda_1$, *the eigen-gap of* $\mathcal{K}_P$. *If the constant* $r$ *defined in* (3.2) *satisfies* $r < t$, *and*

$$(3.3) \qquad \left\| \pi^2 \int_{\mathbb{R}^d} K(x,y) \phi_0^1(y) \, dP^2(y) \right\|_{L_P^2} \leq \varepsilon,$$

*such that* $\varepsilon + r < t$, *then* $\pi^1 \lambda_0^1$ *is close to* $\mathcal{K}_P$'s *top eigenvalue* $\lambda_0$,

$$|\pi^1 \lambda_0^1 - \lambda_0| \leq \varepsilon$$

*and* $\phi_0^1$ *is close to* $\mathcal{K}_P$'s *top eigenfunction* $\phi_0$ *in* $L_P^2$ *sense,*

$$(3.4) \qquad \|\phi_0^1 - \phi_0\|_{L_P^2} \leq \frac{\varepsilon}{t - \varepsilon}.$$

The proof is trivial, so it is omitted here. Since Theorem 3 leads to $|\lambda_0^1 - \lambda_0| \leq r$ and Lemma 1 suggests $|\lambda_0^1 - \lambda_k| \leq \varepsilon$ for some $k$, the condition $r + \varepsilon < t = \lambda_0 - \lambda_1$ guarantees that $\phi_0$ as the only possible choice for $\phi_0^1$ to be close to. Therefore, $\phi_0^1$ is approximately the top eigenfunction of $\mathcal{K}_P$.

It is worth noting that the separable conditions in Theorem 3, Corollary 2 are mainly based on the overlap of the mixture components, but not on their shapes or parametric forms. Therefore, clustering methods based on spectral information are able to deal with more general problems beyond the traditional mixture models based on a parametric family, such as mixture Gaussians or mixture of exponential families.

3.4. *Top spectrum of* $\mathcal{K}_P$ *for mixture distributions.* For a mixture distribution with enough separation between its mixing components, we now extend the perturbation results in Corollary 2 to other top eigenfunctions of $\mathcal{K}_P$. With close agreement between $(\lambda_0, \phi_0)$ and $(\pi^1 \lambda_0^1, \phi_0^1)$, we observe that the second top eigenvalue of $\mathcal{K}_P$ is approximately $\max(\pi^1\lambda_1^1, \pi^2\lambda_0^2)$ by investigating the top eigenvalue of the operator defined by a new kernel $K^{\text{new}} = K(x,y) - \lambda_0\phi_0(x)\phi_0(y)$ and $P$. Accordingly, one may also derive the conditions under which the second eigenfunctions of $\mathcal{K}_P$ is approximated by $\phi_1^1$ or $\phi_0^2$, depending on the magnitude of $\pi^1\lambda_1^1$ and $\pi^2\lambda_0^2$. By sequentially applying the same argument, we arrive at the following corollary.

PROPERTY 1 (Mixture property of top spectrum). *For a convolution operator $\mathcal{K}_P$, defined by a semi-positive definite kernel with a fast tail decay and a mixture distribution $P = \sum_{g=1}^{G} \pi^g P^g$ with enough separations between its mixing components, the top eigenfunctions of $\mathcal{K}_P$ are approximately chosen from the top ones $(\phi_i^g)$ of $\mathcal{K}_{P^g}$, $i = 0, 1, \ldots, n$, $g = 1, \ldots, G$. The ordering of the eigenfunctions is determined by* mixture magnitudes $\pi^g\lambda_i^g$.

This property suggests that each of the top eigenfunctions of $\mathcal{K}_P$ corresponds to exactly one of the separable mixture components. Therefore, we can approximate the top eigenfunctions of $\mathcal{K}_{P^g}$ through those of $\mathcal{K}_P$ when enough separations exist among mixing components. However, several of the top eigenfunctions of $\mathcal{K}_P$ can correspond to the same component and a fixed number of top eigenfunctions may miss some components entirely, specifically the ones with small mixing weights $\pi^g$ or small eigenvalue $\lambda^g$.

When there is a large i.i.d. sample from a mixture distribution whose components are well separated, we expect the top eigenvalues and eigenfunctions of $\mathcal{K}_P$ to be close to those of the empirical operator $\mathcal{K}_{P_n}$. As discussed in Section 2.2, the eigenvalues of $\mathcal{K}_{P_n}$ are the same as those of the kernel matrix $K_n$ and the eigenfunctions of $\mathcal{K}_{P_n}$ coincide with the eigenvectors of $K_n$ on the sampled points. Therefore, assuming good approximation of $\mathcal{K}_{P_n}$ to $\mathcal{K}_P$, the eigenvalues and eigenvectors of $K_n$ provide us with access to the spectrum of $\mathcal{K}_P$.

This understanding sheds light on the algorithms proposed in Scott and Longuet-Higgins [13] and Perona and Freeman [10], in which the top (several) eigenvectors of $K_n$ are used for clustering. While the top eigenvectors may contain clustering information, smaller or less compact groups may not be identified using only the very top part of the spectrum. More eigenvectors need to be investigated to see these clusters. On the other hand, information in the top few eigenvectors may also be redundant for clustering, as some of these eigenvectors may represent the same group.

3.5. *A real-data example: a USPS digits dataset.* Here we use a high-dimensional U.S. Postal Service (USPS) digit dataset to illustrate the properties of the top spectrum of $\mathcal{K}_P$. The data set contains normalized handwritten digits, automatically scanned from envelopes by the USPS. The images here have been rescaled and size-normalized, resulting in $16 \times 16$ grayscale images (see Le Cun et al. [5] for details). Each image is treated as a vector $\mathbf{x}_i$ in $\mathbb{R}^{256}$. In this experiment, 658 "3"s, 652 "4"s and 556 "5"s in the training data are pooled together as our sample (size 1866).

Taking the Gaussian kernel with bandwidth $\omega = 2$, we construct the kernel matrix $K_n$ and compute its eigenvectors $\mathbf{v}_1, \mathbf{v}_2, \ldots, \mathbf{v}_{1866}$. We visualize the digits corresponding to large absolute values of the top eigenvectors. Given an eigenvector $\mathbf{v}_j$, we rank the digits $\mathbf{x}_i$, $i = 1, 2, \ldots, 1866$, according to the absolute value $|(\mathbf{v}_j)_i|$. In each row of Figure 3, we show the 1st, 36th, 71st,...,316th digits according to that order for a fixed eigenvector $\mathbf{v}_j$, $j = 1, 2, 3, 15, 16, 17, 48, 49, 50$. It turns out that the digits with large absolute values of the top 15 eigenvectors, some shown in Figure 3, all represent number "4." The 16th eigenvector is the first one representing "3" and the 49th eigenvector is the first one for "5."

The plot of the data embedded using the top three eigenvectors shown in the left panel of Figure 4 suggests no separation of digits. These results are strongly consistent with our theoretical findings: A fixed number of the top eigenvectors of $K_n$ may correspond to the same cluster while missing other significant clusters. This leads to the failure of clustering algorithms only using the top eigenvectors of $K_n$. The $k$-means algorithm based on top eigenvectors (normalized as suggested in Scott and Longuet-Higgins [13])

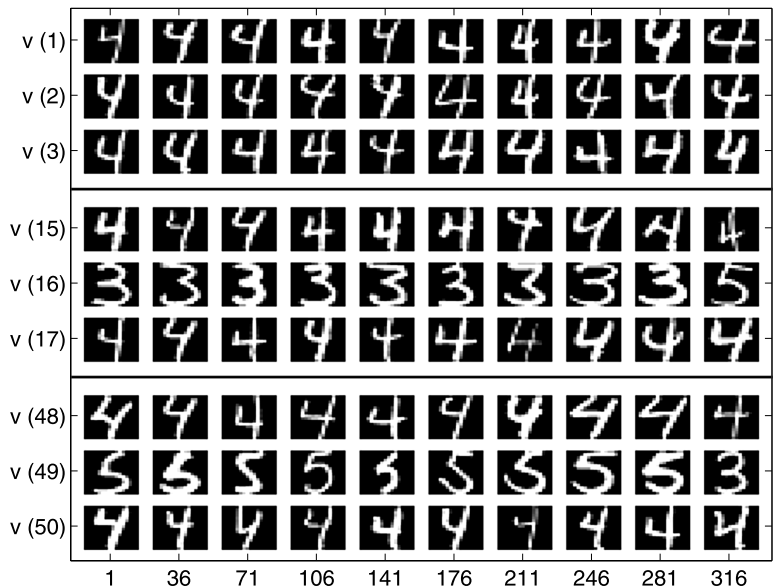

FIG. 3. *Digits ranked by the absolute value of eigenvectors* $\mathbf{v}_1, \mathbf{v}_2, \ldots, \mathbf{v}_{50}$. *The digits in each row correspond to the 1st, 36th, 71st,...,316th largest absolute value of the selected eigenvector. Three eigenvectors,* $\mathbf{v}_1, \mathbf{v}_{16}$ *and* $\mathbf{v}_{49}$, *are identified by our DaSpec algorithm.*

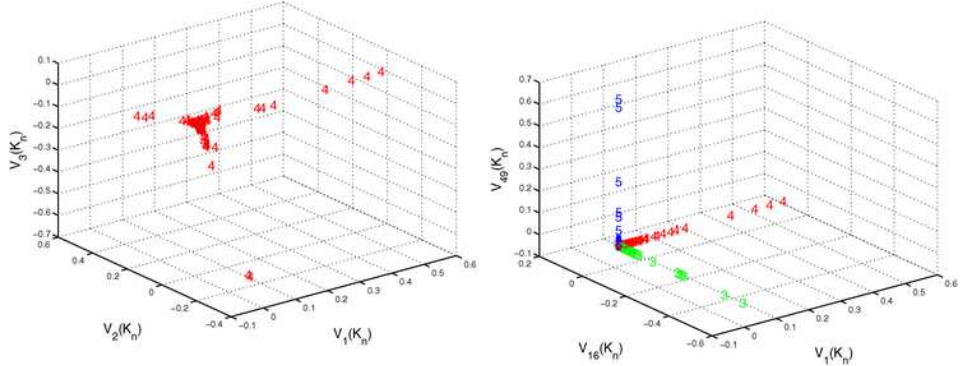

Fig. 4. *Left: scatter plots of digits embedded in the top three eigenvectors; right: digits embedded in the 1st, 16th and 49th eigenvectors.*

produces accuracies below 80% and reaches the best performance as the 49th eigenvector is included.

Meanwhile, the data embedded in the 1st, 16th and 49th eigenvectors (the right panel of Figure 4) do present the three groups of digits "3," "4" and "5" nearly perfectly. If one can intelligently identify these eigenvectors and cluster data in the space spanned by them, good performance is expected. In the next section, we utilize our theoretical analysis to construct a clustering algorithm that automatically selects these most informative eigenvectors and groups the data accordingly.

## 4. A data spectroscopic clustering (DaSpec) algorithm.
In this section, we propose a data spectroscopic clustering (DaSpec) algorithm based on our theoretical analyses. We chose the commonly used Gaussian kernel, but it may be replaced by other positive definite radial kernels with a fast tail decay rate.

### 4.1. Justification and the DaSpec algorithm.
As shown in Property 1 for mixture distributions in Section 3.4, we have access to approximate eigenfunctions of $\mathcal{K}_{P^g}$ through those of $\mathcal{K}_P$ when each mixing component has enough separation from the others. We know from Theorem 2 that among the eigenfunctions of each component $\mathcal{K}_{P^g}$, the top one is the only eigenfunction with no sign change. When the spectrum of $\mathcal{K}_{P^g}$ is close to that of $\mathcal{K}_P$, we expect that there is exactly one eigenfunction with no sign change over a certain small threshold $\varepsilon$. Therefore, the number of separable components of $P$ is indicated by the number of eigenfunctions $\phi(x)$'s of $\mathcal{K}_P$ with no sign change after thresholding.

Meanwhile, the eigenfunctions of each component decay quickly to zero at the tail of its distribution if there is a good separation of components. At a given location $x$, in the high-density area of a particular component which is at the tails of other components, we expect the eigenfunctions from all other components to be close to zero. Among the top eigenfunction $|\phi_0^g(x)|$ of $\mathcal{K}_{P^g}$ defined on each component $p^g$, $g = 1, \ldots, G$, the group identity of $x$ corresponds to the eigenfunction that has the largest absolute value, $|\phi_0^g(x)|$. Combining this observation with previous discussions on the approximation of $K_n$ to $\mathcal{K}_P$, we propose the following clustering algorithm.

*Data spectroscopic clustering (DaSpec) Algorithm.*

     *Input*: Data $x_1, \ldots, x_n \in \mathbb{R}^d$.
*Parameters*: Gaussian kernel bandwidth $\omega > 0$, thresholds $\varepsilon_j > 0$.
    *Output*: Estimated number of separable components $\hat{G}$ and a cluster label $\hat{L}(x_i)$ for each data point $x_i$, $i = 1, \ldots, n$.

STEP 1.  Construct the Gaussian kernel matrix $K_n$:

$$(K_n)_{ij} = \frac{1}{n} e^{-\|x_i - x_j\|^2/(2\omega^2)}, \qquad i, j = 1, \ldots, n,$$

and compute its eigenvalues $\lambda_1, \lambda_2, \ldots, \lambda_n$ and eigenvectors $\mathbf{v}_1, \mathbf{v}_2, \ldots, \mathbf{v}_n$.

STEP 2.  Estimate the number of clusters:

- Identify all eigenvectors $\mathbf{v}_j$ that have no sign changes up to precision $\varepsilon_j$. [We say that a vector $\mathbf{e} = (e_1, \ldots, e_n)'$ has no sign changes up to $\varepsilon$ if either $\forall i \ e_i > -\varepsilon$ or $\forall i \ e_i < \varepsilon$.]
- Estimate the number of groups by $\hat{G}$, the number of such eigenvectors.
- Denote these eigenvectors and the corresponding eigenvalues by $\mathbf{v}_0^1, \mathbf{v}_0^2, \ldots, \mathbf{v}_0^{\hat{G}}$ and $\lambda_0^1, \lambda_0^2, \ldots, \lambda_0^{\hat{G}}$, respectively.

STEP 3.  Assign a cluster label to each data point $x_i$ as

$$\hat{L}(x_i) = \arg\max_g \{|v_{0i}^g| : g = 1, 2, \ldots, \hat{G}\}.$$

It is obviously important to have data-dependent choices for the parameters of the DaSpec algorithm: $\omega$ and $\varepsilon_j$'s. We will discuss some heuristics for those choices in the next section. Given a DaSpec clustering result, one important feature of our algorithm is that little adjustment is needed to classify a new data point $x$. Thanks to the connection between the eigenvector $\mathbf{v}$ of $K_n$ and the eigenfunction $\phi$ of the empirical operator $\mathcal{K}_{P_n}$, we can compute the eigenfunction $\phi_0^g$ corresponding to $\mathbf{v}_0^g$ by

$$\phi_0^g(x) = \frac{1}{\lambda_0^g} \sum_{i=1}^n K(x, x_i) v_{0i}^g, \qquad x \in \mathbb{R}^d.$$

Therefore, Step 3 of the algorithm can be readily applied to any $x$ by replacing $v_{0i}^g$ with $\phi_0^g(x)$. So the algorithm output can serve as a clustering rule that separates not only the data, but also the underline distribution, which is aligned with the motivation behind our Data Spectroscopy algorithm: learning properties of a distribution though the empirical spectrum of $\mathcal{K}_{P_n}$.

4.2. *Data-dependent parameter specification.* Following the justification of our DaSpec algorithm, we provide some heuristics on choosing algorithm parameters in a data-dependent way.

*Gaussian kernel bandwidth $\omega$.* The bandwidth controls both the eigengaps and the tail decay rates of the eigenfunctions. When $\omega$ is too large, the tails of eigenfunctions may not decay fast enough to make condition (3.3) in Corollary 2 hold. However, if $\omega$ is too small, the eigengaps may vanish, in which case each data point will end up as a separate group. Intuitively, we want to select small $\omega$ but still to keep enough (say, $n \times 5\%$) neighbors for most (95% of) data points in the "range" of the kernel, which we define as a length $l$ that makes $P(\|X\| < l) = 95\%$. In case of a Gaussian kernel in $\mathbb{R}^d$, $l = \omega \sqrt{95\% \text{ quantile of } \chi_d^2}$.

Given data $x_1, \ldots, x_n$ or their pairwise $L^2$ distance $d(x_i, x_j)$, we can find $\omega$ that satisfies the above criteria by first calculating $q_i = 5\%$ quantile of $\{d(x_i, x_j), j = 1, \ldots, n\}$ for each $i = 1, \ldots, n$, then taking

$$(4.1) \qquad \omega = \frac{95\% \text{ quantile of } \{q_1, \ldots, q_n\}}{\sqrt{95\% \text{ quantile of } \chi_d^2}}.$$

As shown in the simulation studies in Section 5, this particular choice of $\omega$ works well in low-dimensional case. For high-dimensional data generated from a lower-dimensional structure, such as an $m$-manifold, the procedure usually leads to an $\omega$ that is too small. We suggest starting with $\omega$ defined in (4.1) and trying some neighboring values to see if the results are improved, maybe based on some labeled data, expert opinions, data visualization or trade-off of the between and within cluster distances.

*Threshold $\varepsilon_j$.* When identifying the eigenvectors with no sign changes in Step 2, a threshold $\varepsilon_j$ is included to deal with the small perturbation introduced by other well-separable mixture components. Since $\|\mathbf{v}_j\|^2 = 1$ and the elements of the eigenvector decrease quickly (exponentially) from $\max_i(|\mathbf{v}_j(x_i)|)$, we suggest to threshold $\mathbf{v}_j$ at $\varepsilon_j = \max_i(|\mathbf{v}_j(x_i)|)/n$ ($n$ as the sample size) to accommodate the perturbation.

We note that the proper selection of algorithm parameters is critical to the separation of the spectrum and the success of the clustering algorithms hinged on the separation. Although the described heuristics seem to work well for low-dimensional datasets (as we will show in the next section), they are still preliminary and more research is needed, especially in high-dimensional data analysis. We plan to further study the data-adaptive parameter selection procedure in the future.

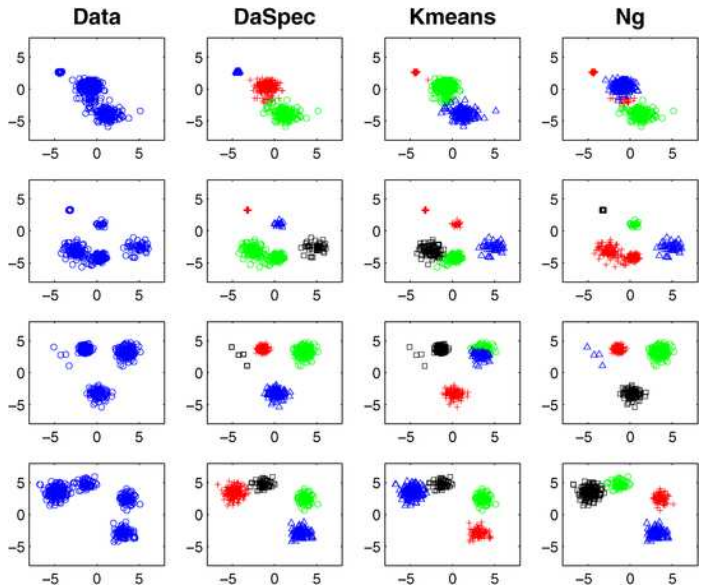

Fig. 5. *Clustering results on four simulated data sets described in Section 5.1. First column: scatter plots of data; second column: results the proposed spectroscopic clustering algorithm; third column: results of the k-means algorithm; fourth column: results of the spectral clustering algorithm (Ng, Jordan and Weiss [8]).*

## 5. Simulation studies.

5.1. *Gaussian type components.* In this simulation, we examine the effectiveness of the proposed DaSpec algorithm on datasets generated from Gaussian mixtures. Each data set (size of 400) is sampled from a mixture of six bivariate Gaussians, while the size of each group follows a Multinomial distribution ($n = 400$, and $p_1 = \cdots = p_6 = 1/6$). The mean and standard deviation of each Gaussian are randomly drawn from a Uniform on $(-5, 5)$ and a Uniform on $(0, 0.8)$, respectively. Four data sets generated from this distribution are plotted in the left column of Figure 5. It is clear that the groups may be highly unbalanced and overlap with each other. Therefore, rather than trying to separate all six components, we expect good clustering algorithms to identify groups with reasonable separations between high density areas.

The DaSpec algorithm is applied with parameters $\omega$ and $\varepsilon_j$ chosen by the procedure described in Section 4.2. Taking the number of groups iden-tified by our Daspec algorithm, the commonly used $k$-means algorithm and the spectral clustering algorithms proposed in Ng, Jordan and Weiss [8] (using the same $\omega$ as the DaSpec) are also tested to serve as baselines for comparison. As a common practice with $k$-means algorithm, fifty random initializations are used and the final results are from the one that minimizes the optimization criterion $\sum_{i=1}^{n}(x_i - y_{k(i)})^2$, where $x_i$ is assigned to group $k(i)$ and $y_k = \sum_{i=1}^{n} x_i I(k(i) = k)/\sum_{i=1}^{n} I(k(i) = k)$.

As shown in the second column of Figure 5, the proposed DaSpec algorithm (with data-dependent parameter choices) identifies the number of separable groups, isolates potential outliers and groups data accordingly. The results are similar to the $k$-means algorithm results (the third column), when the groups are balanced and their shapes are close to round. In these cases, the $k$-means algorithm is expected to work well, given that the data in each group is well represented by its average. The last column shows the results of Ng et al.'s spectral clustering algorithm, which sometimes (see the first row) assigns data to one group even when they are actually far away.

In summary, for this simulated example, we find that the proposed DaSpec algorithm, with data-adaptively chosen parameters, identifies the number of separable groups reasonably well and produces good clustering results when the separations are large enough. It is also interesting to note that the algorithm isolates possible "outliers" into a separate group so that they do not affect the clustering results on the majority of data. The proposed algorithm competes well against the commonly used $k$-means and spectral clustering algorithms.

5.2. *Beyond Gaussian components.* We now compare the performance of the aforementioned clustering algorithms on data sets that contain non-Gaussian groups, various levels of noise and possible outliers. Data set $\mathcal{D}_1$ contains three well-separable groups and an outlier in $\mathbb{R}^2$. The first group of data is generated by adding independent Gaussian noise $N((0,0)^T, 0.15^2 I_{2\times 2})$ to 200 uniform samples from three fourths of a ring with radius 3, which is from the same distribution as those plotted in the right panel of Figure 8. The second group includes 100 data points sampled from a bivariate Gaussian $N((3,-3)^T, 0.5^2 I_{2\times 2})$, and the last group has only 5 data points sampled from a bivariate Gaussian $N((0,0)^T, 0.3^2 I_{2\times 2})$. Finally, one outlier is located at $(5,5)^T$. Given $\mathcal{D}_1$, three more data sets ($\mathcal{D}_2$, $\mathcal{D}_3$ and $\mathcal{D}_4$) are created by gradually adding independent Gaussian noise (with standard deviations 0.3, 0.6, 0.9, respectively). The scatter plots of the four datasets are shown in the left column of Figure 6. It is clear that the degree of separation decreases from top to bottom.

Similarly to the previous simulation, we examine the DaSpec algorithm with data-driven parameters, the $k$-means and Ng et al.'s spectral clustering algorithms on these data sets. The latter two algorithms are tested under two different assumptions on the number of groups: the number $(G)$ identified by the DaSpec algorithm or one group less $(G − 1)$. Note that the DaSpec algorithm claims only one group for $\mathcal{D}_4$, so the other two algorithms are not applied to $\mathcal{D}_4$.

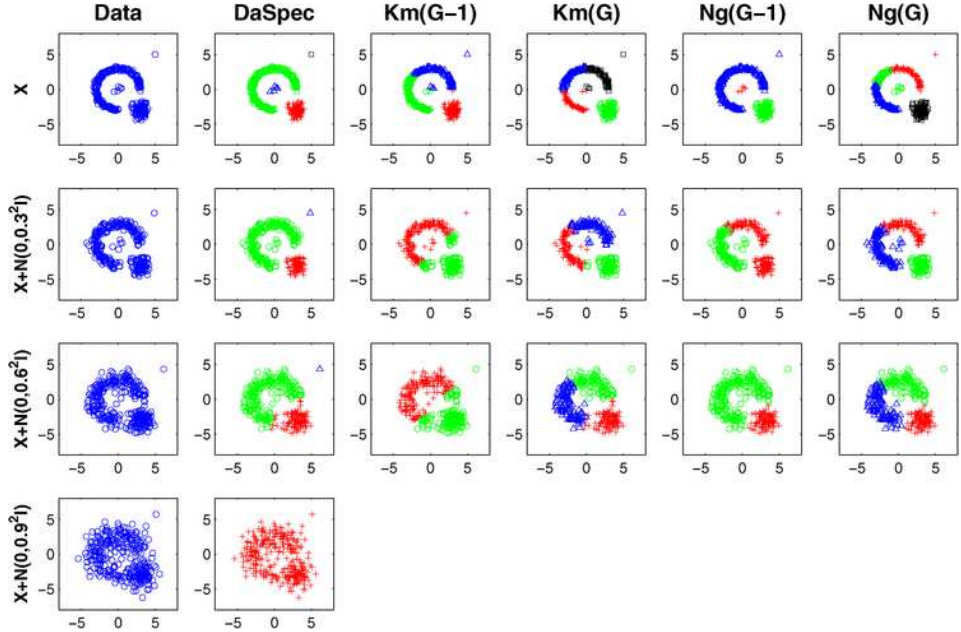

FIG. 6. *Clustering results on four simulated data sets described in Section 5.2. First column: scatter plots of data; second column: labels of the G identified groups by the proposed spectroscopic clustering algorithm; third and forth columns: k-means algorithm assuming G − 1 and G groups, respectively; fifth and sixth columns: spectral clustering algorithm (Ng, Jordan and Weiss [8]) assuming G − 1 and G groups, respectively.*

The DaSpec algorithm (the second column in the right panel of Figure 6) produces a reasonable number of groups and clustering results. For the perfectly separable case in $\mathcal{D}_1$, three groups are identified and the one outlier is isolated out. It is worth noting that the incomplete ring is separated from other groups, which is not a simple task for algorithms based on group centroids. We also see that the DaSpec algorithm starts to combine inseparable groups as the components become less separable.

Not surprisingly, the $k$-means algorithms (the third and fourth columns) do not perform well because of the presence of the non-Gaussian component, unbalanced groups and outliers. Given enough separations, the spectral clustering algorithm reports reasonable results (the fifth and sixth columns). However, it is sensitive to outliers and the specification of the number of groups.

**6. Conclusions and discussion.** Motivated by recent developments in kernel and spectral methods, we study the connection between a probability distribution and the associated convolution operator. For a convolution operator defined by a radial kernel with a fast tail decay, we show that each top eigenfunction of the convolution operator defined by a mixture distribution is approximated by one of the top eigenfunctions of the operator corresponding to a mixture component. The separation condition is mainly based on the overlap between high-density components, instead of their explicit parametric forms, and thus is quite general. These theoretical results explain why the top eigenvectors of kernel matrix may reveal the clustering information but do not always do so. More importantly, our results reveal that not every component will contribute to the top few eigenfunctions of the convolution operator $\mathcal{K}_P$ because the size and configuration of a component decides the corresponding eigenvalues. Hence the top eigenvectors of the kernel matrix may or may not preserve all clustering information, which explains some empirical observations of certain spectral clustering methods.

Following our theoretical analyses, we propose the data spectroscopic clustering algorithm based on finding eigenvectors with no sign change. Comparing to commonly used $k$-means and spectral clustering algorithms, DaSpec is simple to implement and provides a natural estimator of the number of separable components. We found that DaSpec handles unbalanced groups and outliers better than the competing algorithms. Importantly, unlike $k$-means and certain spectral clustering algorithms, DaSpec does not require random initialization, which is a potentially significant advantage in practice. Simulations show favorable results compared to $k$-means and spectral clustering algorithms. For practical applications, we also provide some guidelines for choosing the algorithm parameters.

Our analyses and discussions on connections to other spectral or kernel methods shed light on why radial kernels, such as Gaussian kernels, perform well in many classification and clustering algorithms. We expect that this line of investigation would also prove fruitful in understanding other kernel algorithms, such as Support Vector Machines.

## APPENDIX A

Here we provide three concrete examples to illustrate the properties of the eigenfunction of $\mathcal{K}_P$ shown in Section 3.1.

EXAMPLE 1 (Gaussian kernel, Gaussian density). Let us start with the univariate Gaussian case where the distribution $P \sim N(\mu, \sigma^2)$ and the kernel function is also Gaussian. Shi, Belkin and Yu [15] provided the eigenvalues and eigenfunctions of $\mathcal{K}_P$, and the result is a slightly refined version of a result in Zhu et al. [22].

Proposition 1. *For $P \sim N(\mu, \sigma^2)$ and a Gaussian kernel $K(x, y) = e^{-(x-y)^2/(2\omega^2)}$, let $\beta = 2\sigma^2/\omega^2$ and let $H_i(x)$ be the ith order Hermite polynomial. Then eigenvalues and eigenfunctions of $\mathcal{K}_P$ for $i = 0, 1, \ldots$ are given by*

$$\lambda_i = \sqrt{\frac{2}{(1 + \beta + \sqrt{1 + 2\beta})}} \left( \frac{\beta}{1 + \beta + \sqrt{1 + 2\beta}} \right)^i,$$

$$\phi_i(x) = \frac{(1 + 2\beta)^{1/8}}{\sqrt{2^i i!}} \exp \left( -\frac{(x - \mu)^2}{2\sigma^2} \frac{\sqrt{1 + 2\beta} - 1}{2} \right) H_i \left( \left( \frac{1}{4} + \frac{\beta}{2} \right)^{1/4} \frac{x - \mu}{\sigma} \right).$$

Here $H_k$ is the $k$th order Hermite polynomial. Clearly from the explicit expression and expected from Theorem 2, $\phi_0$ is the only positive eigenfunction of $\mathcal{K}_P$. We note that each eigenfunction $\phi_i$ decays quickly (as it is a Gaussian multiplied by a polynomial) away from the mean $\mu$ of the probability distribution. We also see that the eigenvalues of $\mathcal{K}_P$ decay exponentially with the rate dependent on the bandwidth of the Gaussian kernel $\omega$ and the variance of the probability distribution $\sigma^2$. These observations can be easily generalized to the multivariate case; see Shi, Belkin and Yu [15].

Example 2 (Exponential kernel, uniform distribution on an interval). To give another concrete example, consider the exponential kernel $K(x, y) = \exp(-\frac{|x-y|}{\omega})$ for the uniform distribution on the interval $[-1, 1] \subset \mathbb{R}$. In Diaconis, Goel and Holmes [3] it was shown that the eigenfunctions of this kernel can be written as $\cos(bx)$ or $\sin(bx)$ inside the interval $[-1, 1]$ for appropriately chosen values of $b$ and decay exponentially away from it. The top eigenfunction can be written explicitly as follows:

$$\phi(x) = \frac{1}{\lambda} \int_{[-1,1]} e^{-|x-y|/\omega} \cos(by) \, dy \qquad \forall x \in \mathbb{R},$$

where $\lambda$ is the corresponding eigenvalue. Figure 7 illustrates an example of this behavior, for $\omega = 0.5$.

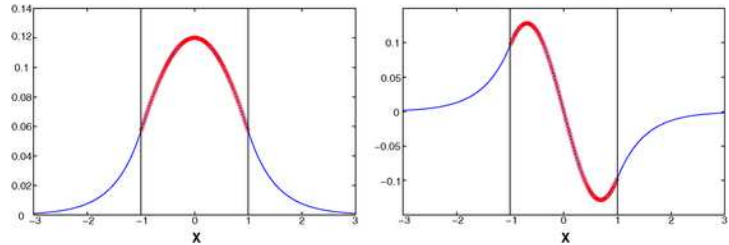

Fig. 7.    *Top two eigenfunctions of the exponential kernel with bandwidth $\omega = 0.5$ and the uniform distribution on $[-1, 1]$.*

EXAMPLE 3 (A curve in $\mathbb{R}^d$). We now give a brief informal discussion of the important case when our probability distribution is concentrated on or around a low-dimensional submanifold in a (potentially high dimensional) ambient space. The simplest example of this setting is a Gaussian distribution, which can be viewed as a zero-dimensional manifold (the mean of the distribution) plus noise.

A more interesting example of a manifold is a curve in $\mathbb{R}^d$. We observe that such data is generated by any time-dependent smooth deterministic process, whose parameters depend continuously on time $t$. Let $\psi(t):[0,1] \to \mathbb{R}^d$ be such a curve. Consider a restriction of the kernel $\mathcal{K}_P$ to $\psi$. Let $x,y \in \psi$ and let $d(x,y)$ be the geodesic distance along the curve. It can be shown that $d(x,y) = \|x-y\| + O(\|x-y\|^3)$, when $x,y$ are close, with the remainder term depending on how the curve is embedded in $\mathbb{R}^d$. Therefore, we see that if the kernel $\mathcal{K}_P$ is a sufficiently local radial basis kernel, the restriction of $\mathcal{K}_P$ to $\psi$ is a perturbation of $\mathcal{K}_P$ in a one-dimensional case. For the exponential kernel, the one-dimensional kernel can be written explicitly (see Example 2), and we have an approximation to the kernel on the manifold with a decay off the manifold (assuming that the kernel is a decreasing function of the distance). For the Gaussian kernel, a similar extension holds, although no explicit formula can be easily obtained.

The behaviors of the top eigenfunction of the Gaussian and exponential kernel, respectively, are demonstrated in Figure 8. The exponential kernel corresponds to the bottom left panel. The behavior of the eigenfunction is seen generally consistent with the top eigenfunction of the exponential kernel on $[-1,1]$ shown in Figure 8. The Gaussian kernel (top left panel) has similar behaviors but produces level lines more consistent with the data distribution, which may be preferable in practice. Finally, we observe that the addition of small noise (right top and bottom panels) does not significantly change the eigenfunctions.

## APPENDIX B

PROOF OF THEOREM 2. For a semi-positive definite kernel $K(x,y)$ with full support on $\mathbb{R}^d$, we first show the top eigenfunction $\phi_0$ of $\mathcal{K}_P$ has no sign change on the support of the distribution. We define $R^+ = \{x \in \mathbb{R}^d : \phi_0(x) > 0\}$, $R^- = \{x \in \mathbb{R}^d : \phi_0(x) < 0\}$ and $\bar{\phi}_0(x) = |\phi_0(x)|$. It is clear that $\int \bar{\phi}_0^2 \, dP = \int \phi_0^2 \, dP = 1$.

Assuming that $P(R^+) > 0$ and $P(R^-) > 0$, we will show that

$$\iint K(x,y)\bar{\phi}_0(x)\bar{\phi}_0(y)\,dP(x)\,dP(y)$$

$$> \iint K(x,y)\phi_0(x)\phi_0(y)\,dP(x)\,dP(y),$$

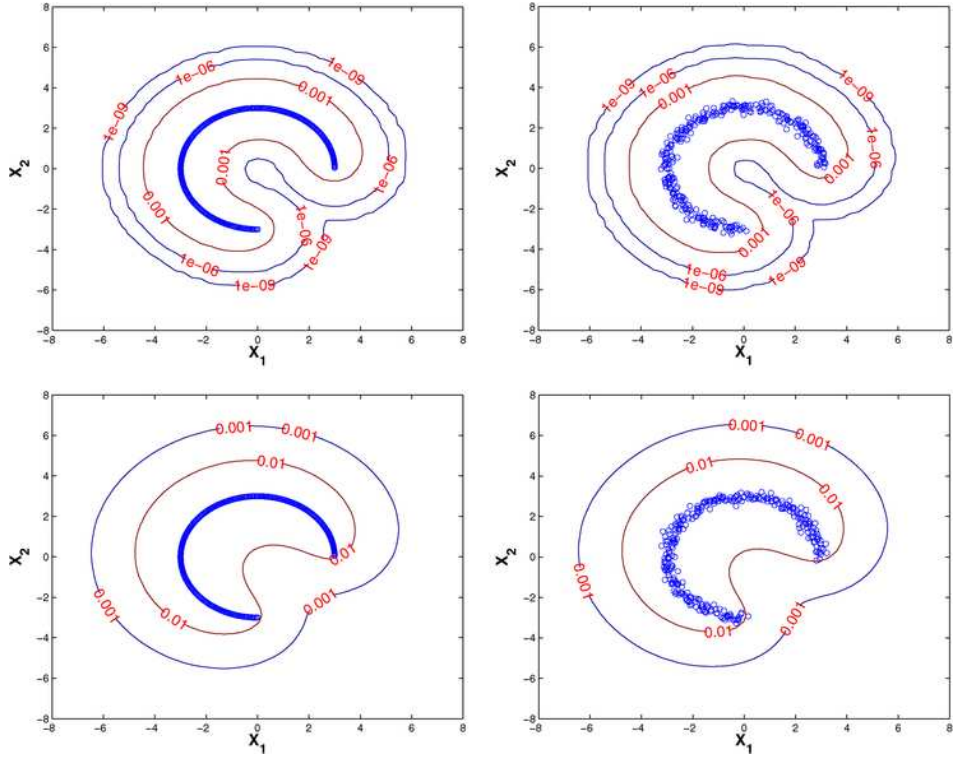

Fɪɢ. 8.   *Contours of the top eigenfunction of $\mathcal{K}_P$ for Gaussian (upper panels) and exponential kernels (lower panels) with bandwidth 0.7. The curve is 3/4 of a ring with radius 3 and independent noise of standard deviation 0.15 added in the right panels.*

which contradicts with the assumption that $\phi_0(\cdot)$ is the eigenfunction associated with the largest eigenvalue. Denoting $g(x, y) = K(x, y)\phi_0(x)\phi_0(y)$ and $\bar{g}(x, y) = K(x, y)\bar{\phi}_0(x)\bar{\phi}_0(y)$, we have

$$\int_{R^+} \int_{R^+} \bar{g}(x, y)\, dP(x)\, dP(y)$$
$$= \int_{R^+} \int_{R^+} g(x, y)\, dP(x)\, dP(y)$$

and the equation also holds on region $R^- \times R^-$. However, over the region $\{(x, y) : x \in R^+ \text{ and } y \in R^-\}$, we have

$$\int_{R^+} \int_{R^-} \bar{g}(x, y)\, dP(x)\, dP(y)$$
$$> \int_{R^+} \int_{R^-} g(x, y)\, dP(x)\, dP(y),$$

since $K(x,y) > 0$, $\phi_0(x) > 0$ and $\phi_0(y) < 0$. The inequality holds on $\{(x,y) : x \in R^-$ and $y \in R^+\}$. Putting four integration regions together, we arrive at the contradiction. Therefore, the assumptions $P(R^+) > 0$ and $P(R^-) > 0$ cannot be true at the same time, which implies that $\phi_0(\cdot)$ has no sign changes on the support of the distribution.

Now consider $\forall x \in \mathbb{R}^d$. We have

$$\lambda_0 \phi_0(x) = \int K(x,y)\phi_0(y) \, dP(y).$$

Given the facts that $\lambda_0 > 0$, $K(x,y) > 0$, and $\phi_0(y)$ have the same sign on the support, it is straightforward to see that $\phi_0(x)$ has no sign changes and has full support in $\mathbb{R}^d$. Finally, the isolation of $(\lambda_0, \phi_0)$ follows. If there exist another $\phi$ that shares the same eigenvalue $\lambda_0$ with $\phi_0$, they both have no sign change and have full support on $\mathbb{R}^d$. Therefore, $\int \phi_0(x)\phi(x) \, dP(x) > 0$ and it contradicts with the orthogonality between eigenfunctions. $\quad\square$

PROOF OF THEOREM 3. By definition, the top eigenvalue of $\mathcal{K}_P$ satisfies

$$\lambda_0 = \max_f \frac{\iint K(x,y)f(x)f(y) \, dP(x) \, dP(y)}{\int [f(x)]^2 \, dP(x)}.$$

For any function $f$,

$$\iint K(x,y)f(x)f(y) \, dP(x) \, dP(y)$$

$$= [\pi^1]^2 \iint K(x,y)f(x)f(y) \, dP^1(x) \, dP^1(y)$$

$$+ [\pi^2]^2 \iint K(x,y)f(x)f(y) \, dP^2(x) \, dP^2(y)$$

$$+ 2\pi^1 \pi^2 \iint K(x,y)f(x)f(y) \, dP^1(x) \, dP^2(y)$$

$$\leq [\pi^1]^2 \lambda_0^1 \int [f(x)]^2 \, dP^1(x) + [\pi^2]^2 \lambda_0^2 \int [f(x)]^2 \, dP^2(x)$$

$$+ 2\pi^1 \pi^2 \iint K(x,y)f(x)f(y) \, dP^1(x) \, dP(y)^2.$$

Now we concentrate on the last term,

$$2\pi^1 \pi^2 \iint K(x,y)f(x)f(y) \, dP^1(x) \, dP^2(y)$$

$$\leq 2\pi^1 \pi^2 \sqrt{\iint [K(x,y)]^2 \, dP^1(x) \, dP^2(y)}$$

$$\times \sqrt{\iint [f(x)]^2 [f(y)]^2 \, dP^1(x) \, dP^2(y)}$$

$$= 2\sqrt{\pi^1 \pi^2 \iint [K(x,y)]^2 \, dP^1(x) \, dP^2(y)}$$

$$\times \sqrt{\pi^1 \int [f(x)]^2 \, dP^1(x)} \sqrt{\pi^2 \int [f(y)]^2 \, dP^2(y)}$$

$$\leq \sqrt{\pi^1 \pi^2 \iint [K(x,y)]^2 \, dP^1(x) \, dP^2(y)}$$

$$\times \left( \pi^1 \int [f(x)]^2 \, dP^1(x) + \pi^2 \int [f(x)]^2 \, dP^2(x) \right)$$

$$= r \int [f(x)]^2 \, dP(x),$$

where $r = (\pi^1 \pi^2 \iint [K(x,y)]^2 \, dP^1(x) \, dP^2(y))^{1/2}$. Thus,

$$\lambda_0 = \max_{f \,:\, \int f^2 \, dP = 1} \iint K(x,y) f(x) f(y) \, dP(x) \, dP(y)$$

$$\leq \max_{f \,:\, \int f^2 \, dP = 1} \left[ \pi^1 \lambda_0^1 \int [f(x)]^2 \pi^1 \, dP^1(x) + \pi^2 \lambda_0^2 \int [f(x)]^2 \pi^2 \, dd P^2(x) + r \right]$$

$$\leq \max(\pi^1 \lambda_0^1, \pi^2 \lambda_0^2) + r.$$

The other side of the equality is easier to prove. Assuming $\pi^1 \lambda_0^1 > \pi^2 \lambda_0^2$ and taking the top eigenfunction $\phi_0^1$ of $\mathcal{K}_{P^1}$ as $f$, we derive the following results by using the same decomposition on $\iint K(x,y) \phi_0^1(x) \phi_0^1(y) \, dP(x) \, dP(y)$ and the facts that $\int K(x,y) \phi_0^1(x) \, dd P^1(x) = \lambda_0^1 \phi_0^1(y)$ and $\int [\phi_0^1]^2 \, dP^1 = 1$. Denoting $h(x,y) = K(x,y) \phi_0^1(x) \phi_0^1(y)$, we have

$$\lambda_0 \geq \frac{\iint K(x,y) \phi_0^1(x) \phi_0^1(y) \, dP(x) \, dP(y)}{\int [\phi_0^1(x)]^2 \, dP(x)}$$

$$= \frac{[\pi^1]^2 \lambda_0^1 + [\pi^2]^2 \iint h(x,y) \, dP^2(x) \, dP^2(y) + 2\pi^1 \pi^2 \lambda_0^1 \int [\phi_0^1(x)]^2 \, dP^2(x)}{\pi^1 + \pi^2 \int [\phi_0^1(x)]^2 \, dP^2(x)}$$

$$= \pi^1 \lambda_0^1 \left( \frac{\pi^1 + 2\pi^2 \int [\phi_0^1(x)]^2 \, dP^2(x)}{\pi^1 + \pi^2 \int [\phi_0^1(x)]^2 \, dP^2(x)} \right) + \frac{[\pi^2]^2 \iint h(x,y) \, dP^2(x) \, dP^2(y)}{\pi^1 + \pi^2 \int [\phi_0^1(x)]^2 \, dP^2(x)}$$

$$\geq \pi^1 \lambda_0^1.$$

This completes the proof. □

**Acknowledgment.** The authors would like to thank Yoonkyung Lee, Prem Goel, Joseph Verducci and Donghui Yan for helpful discussions, suggestions and comments.

T. Shi
Department of Statistics
Ohio State University
1958 Neil Avenue, Cockins Hall 404
Columbus, Ohio 43210-1247
USA
E-mail: taoshi@stat.osu.edu

M. Belkin
Department of Computer Science
  and Engineering
Ohio State University
2015 Neil Avenue, Dreese Labs 597
Columbus, Ohio 43210-1277
USA
E-mail: mbelkin@sce.osu.edu

B. Yu
Department of Statistics
University of California, Berkeley
367 Evans Hall
Berkeley, California 94720-3860
USA
E-mail: binyu@stat.berkeley.edu